\documentclass[conference]{IEEEtran_1}
\IEEEoverridecommandlockouts
\usepackage{multirow}
\usepackage{booktabs}
\usepackage[hidelinks]{hyperref}
\usepackage{cite}
\usepackage{makecell}
\usepackage{amsmath,amssymb,amsfonts}
\usepackage{algorithmic}
\usepackage{graphicx}
\usepackage{textcomp}
\usepackage{xcolor}

\begin{document}

\title{\vspace{12pt}HiLLTS: Zero-Shot Hierarchical LLM-Guided Traffic Signal Control for Sustainable Transportation}


\author{Yue Ding, Tendai Mukande and Mingming Liu  \thanks{Y. Ding is with the Research Ireland Centre for Researching Training in Machine Learning at Dublin City University. T. Mukande and M. Liu are with the Insight Research Ireland Centre for Data Analytics and the School of Electronic Engineering, Dublin City University. This research is supported by \textit{Taighde Éireann} - Research Ireland under Grant \textit{SFI/12/RC/2289\_P2 }(Insight Research Ireland Centre for Data Analytics) and  Grant \textit{18/CRT/6183} Research Ireland Centre for Research Training in Machine Learning (ML-LABS). \textit{Corresponding author: Mingming Liu. Email: {\tt mingming.liu@dcu.ie}.}}}


\maketitle

\begin{abstract}
Urban traffic congestion significantly increases fuel consumption, greenhouse gas emissions, and commuter delays, resulting in substantial economic losses and environmental harm in modern cities. Traditional traffic signal control strategies such as fixed-time scheduling, actuated control, and reinforcement learning (RL)-based methods, offer different degrees of adaptability; however, RL-based methods can require extensive retraining, careful reward design, and substantial simulation data when transferred across networks or demand regimes. To address these challenges, we propose HiLLTS, an LLM-guided traffic signal control framework that employs a hierarchical three-layer architecture consisting of a central coordination agent, a district layer and multiple cluster-level intersection agents. Experimental results demonstrate consistent improvements in both 
congestion and environmental performance. Compared with the strongest 
non-LLM baseline in each scenario, HiLLTS reduces average waiting 
time by 36.73\% under the low-congestion scenario and 14.71\% under 
the high-congestion scenario, while reducing average CO$_2$ emissions 
by 7.87\% and 8.57\%, respectively. Larger gains are observed against 
weaker baselines: under low congestion, HiLLTS achieves reductions of 
up to 18.00\% in emissions and 62.07\% in waiting time relative to 
Fixed-Time control; under high congestion, reductions of up to 28.89\% 
in emissions and 40.36\% in waiting time are observed relative to Max 
Pressure. The ablation study further validates the contribution of 
LLM-guided coordination over rule-based control.

\end{abstract}


\section{Introduction}
Urban transportation systems around the world are increasingly strained by population growth and increasing vehicle ownership. As traffic demand approaches or exceeds infrastructure capacity, cities experience persistent congestion, with longer travel times resulting in higher carbon dioxide emissions. These impacts impose substantial economic costs through productivity losses and wasted energy, while also contributing to environmental degradation and public health concerns \cite{khreis2017health}. Inefficient signal timing can lead to queue spills, stop-and-go driving, and increased emissions. In contrast, adaptive coordination strategies can significantly reduce delays and improve overall network efficiency without requiring additional infrastructure \cite{eom2020traffic}.

To improve adaptability in dynamic traffic conditions, AI models such as RL have been adopted to optimise signal control policies \cite{chu2019multi}. By learning from real-time state transitions and reward feedback, RL-based methods enable data-driven and context-aware decision-making. More recently, large language models (LLMs) have also emerged as a promising paradigm in intelligent transportation systems, offering advanced reasoning capabilities and processing heterogeneous information \cite{choi2025optimizing}. This shift opens new opportunities for integrating automated reasoning with traffic control, potentially overcoming the limitations of existing methods. For example, \cite{tits2025crossroads} applies LLMs to offline signal optimization, while \cite{yan2025llm} and \cite{vehicles2025} validate real-time control at single intersections. Extensions to grid-based multi-intersection networks are presented in \cite{llmlight2025, llmrl2025}. Despite these advances, several limitations remain. Firstly, most existing studies evaluate their methods in simulated grid networks that do not capture the complexity of roads in the real-world, thus limiting the generalisability of their findings \cite{anthi2025role}. In addition, previous research has focused primarily on reducing congestion and minimising travel delays. However, the environmental impact of such driving patterns has received relatively limited attention \cite{wang2023carbon}. 

To address these challenges, we propose HiLLTS, an LLM-guided hierarchical signal control framework that integrates global traffic coordination with local responsiveness, without requiring costly infrastructure modifications. We argue that reducing carbon emissions is imperative given the increasing global emphasis on sustainable urban mobility and climate-conscious transportation planning. Therefore, HiLLTS is designed to operate efficiently without retraining the LLM, minimising additional computational overhead and associated environmental impact. HiLLTS adopts a three-layer architecture. At the city level, an LLM-guided agent periodically analyzes traffic conditions throughout the network and issues coordination directives to each district. At the district level, cluster metrics are aggregated, and city-level parameters are broadcast to all clusters within the district. At the cluster level, each cluster controller executes actuated gap-detection at every simulation step while overlaying city-directed coordination. The framework is evaluated on the Dublin City Centre (DCC) map, comprising 434 traffic lights organised into 5 districts and 176 clusters. The topology reflects the structural complexity of real urban traffic systems, more structurally realistic than synthetic grid networks. 
 Experimental results demonstrate that HiLLTS reduces both travel time and carbon dioxide emissions, compared to the three baseline models. Although the results are promising, the evaluation remains simulation-based and further validation is required before drawing conclusions about real-world deployment. Our main \textbf{contributions} are summarised as follows:

\begin{itemize}
    \item We propose a novel hierarchical LLM-guided architecture 
    for traffic signal control that requires no training.
  
    \item To the best of our knowledge, this is among the first studies to evaluate zero-shot LLM-guided traffic signal coordination on an irregular city-scale SUMO network derived from real urban topology, showing consistent improvements in congestion mitigation and emission reduction over strong adaptive baselines.
\end{itemize}

The remainder of this paper is organized as follows. Section \ref{sec:litRev} reviews related work on LLM-based traffic signal control and carbon-aware traffic optimization. Section \ref{sec:problem} formulates the hierarchical traffic control problem. Section \ref{sec:architecture} presents the proposed HiLLTS framework. Section \ref{sec:experiments} describes the experimental setup. Section \ref{sec:results}
presents the results and ablation study. Section \ref{sec:conclusion} concludes the paper.

\section{Related Work}
\label{sec:litRev}

In this section, we review relevant work on LLM-based traffic signal control, progressing from smaller to larger experimental scales, and survey work related to carbon emission awareness in traffic control. 

\subsection{Single Intersection LLM-Based Traffic Signal Control}

Recent advances in LLMs have opened new avenues for adaptive 
traffic signal control. Early work demonstrated the feasibility of deploying LLMs as single-intersection controllers.  \cite{yan2025llm} introduced LLM-TrafficBrain, an information-centric framework that integrates semantic reasoning with a closed-loop feedback mechanism. Under  Simulation of Urban MObility (SUMO), it demonstrated superior efficiency, reducing average delay and queue length compared to heuristic adaptive methods. \cite{vehicles2025} further explored chain-of-thought (CoT) prompting with multiple LLM backends (GPT-4o-mini, Gemini, Llama) at a single four-leg intersection. \cite{tits2025crossroads} conducted a systematic comparison between zero-shot CoT and a novel Generative Critic Agent architecture in SUMO. This actor–critic text-feedback approach demonstrated that structured semantic feedback can reduce delay and increase average speed compared to traditional Fixed-Time control.

\subsection{Multi-Intersection and Network-Wide LLM Control}

Scaling LLM-based control from single intersections to city-scale road networks
introduces significant coordination challenges. \cite{llmlight2025} proposed LLMLight, which formulates traffic signal control as a partially observable Markov Game and employs commonsense-augmented step-by-step reasoning to select signal phases. The scalability of the framework was validated on grid-style urban road networks of varying complexity: from smaller grids in Jinan and Hangzhou to a dense large-scale grid in New York with 196 intersections. By effectively managing these structured networks, LLMLight achieved significant reductions in average travel time, average waiting time, and queue length. Similarly, \cite{llmrl2025} proposed an LLM-based RL hybrid framework that predicts congestion bottlenecks, resulting in a reduction in average delay and an increase in throughput using the same map as LLMLight. \cite{icca2025} proposed an LLM-enhanced Transformer-MAPPO framework across 4–6 junctions, improving average reward and convergence speed over the MAPPO baseline.

\subsection{Carbon Emission Awareness in Traffic Optimization}

Despite significant progress in LLM-based signal control, carbon emission reduction remains an underexplored objective. Existing work primarily optimises throughput, average travel time, and queue length, with only implicit environmental benefits arising from reduced idling. \cite{hybridtrafficai2025} proposed a city-level generative AI framework that combines multi-modal sensor fusion with LLM semantic reasoning for vehicle behavior modeling, achieving a normalised risk score of 0.910, but stops short of direct emission quantification. \cite{matsim2025} integrated LLM-based replanning agents into MATSim to optimise electric vehicle charging in Montreal, demonstrating the applicability of LLM in emission-sensitive mobility scenarios. A comprehensive survey by \cite{survey2025} further confirms that carbon-aware traffic signal control remains an open problem, with most LLM-based methods lacking explicit emission feedback mechanisms. This gap motivates the design of our emission-aware  formulation and real-world SUMO-based evaluation.




\section{Problem Formulation}
\label{sec:problem}


We model the urban road network as a directed graph $\mathcal{G} = (\mathcal{V}, \mathcal{E})$, where $\mathcal{V}$ denotes the set of intersections and $\mathcal{E}$ denotes the set of road segments connecting them. A subset $\mathcal{L} \subseteq \mathcal{V}$ represents signalised intersections, which are grouped into clusters $\mathcal{C}$, where each cluster is managed by a cluster controller. The full network is organised into a three-level hierarchy: city-level, district-level, and cluster-level which is constructed through the following three-stage pipeline:

\textit{1) Geographic District Partitioning:} The $|\mathcal{L}|$ signalised intersections are partitioned into $K$ geographic districts. Each intersection $l_i \in \mathcal{L}$ has a geographic coordinate $(x_i, y_i)$. Using $K$-means clustering \cite{mcqueen1967some}, each intersection is assigned to the district with the nearest centroid. 
The district index $k^*(i)$ assigned to intersection $l_i$ is:

\begin{equation}
    k^*(l_i) = \arg\min_{k \in \{1,\ldots,K\}} 
    \left\| (x_i, y_i) - \boldsymbol{\mu}_k \right\|^2
\end{equation}

\noindent where $\boldsymbol{\mu}_k \in \mathbb{R}^2$ is the centroid of district $k$, computed as the mean coordinate of all intersections assigned to it. This yields a partition $\{\mathcal{D}_1, \mathcal{D}_2, \ldots, \mathcal{D}_K\}$ of $\mathcal{L}$, where each district $\mathcal{D}_k \subseteq \mathcal{L}$ groups geographically proximate intersections.

\textit{2) Topological Cluster Partitioning:} Within each district $\mathcal{D}_k$, a road connectivity graph $G_k = (V_k, E_k)$ is constructed, where nodes are signalised intersections and edges are direct road links between them, weighted by the number of lanes. The Louvain community detection algorithm \cite{blondel2008fast} is then applied to $G_k$ to identify fine-grained clusters $\{\mathcal{C}_{k,1}, \mathcal{C}_{k,2}, \ldots\}$, maximising modularity:
\begin{equation}
    Q = \frac{1}{2|E_k|} \sum_{i,j} \left[ w_{ij} - \frac{d_i d_j}{2|E_k|} \right] \delta(c_i, c_j)
\end{equation}
where $w_{ij}$ is the edge weight (lane count), and $d_i = \sum_{j} w_{ij}$ 
is the weighted degree of intersection $i$, and $\delta(c_i, c_j) = 1$ if intersections $i$ and $j$ belong to the same community. This ensures that clusters reflect genuine topological connectivity rather than purely spatial proximity.

\textit{3) District and City Topology Aggregation:} District-level topology is built by merging the connectivity of all clusters within a district. Connections between clusters are labelled as \textit{internal} if both endpoints belong to the same district, or \textit{external} if they cross district boundaries. External links are used to identify adjacent districts. At the city level, all inter-district connections are aggregated into a single graph, and districts in the top third by bottleneck ratio are marked as \textit{core districts}, indicating areas with the most structural congestion risk. A \textit{bottleneck} is an intersection with limited outflow capacity. A cluster is designated a bottleneck cluster if it contains at least one such intersection. This city-level graph is passed to the LLM agent for global coordination decisions.

\subsection{Hierarchical Decision Problem}

The control objective is to improve overall network 
performance in terms of congestion and environmental 
impact. Rather than optimising a fixed mathematical 
objective, the city-level LLM agent is instructed to 
minimise vehicle delay and associated emissions across 
the urban road network, and determines coordination 
directives through structured natural language reasoning 
based on observed district conditions. The overall control problem is decomposed into three hierarchical levels. At the city level, at time \textit{t}, an LLM-guided agent $\pi_{\text{city}}$ observes aggregated district-level statistics $\mathbf{s}_d(t)$, including weighted average waiting time and total CO$_2$ and produces coordination directives $\mathbf{a}_d(t)$ for each district:
\begin{equation}
    \mathbf{a}_d(t) = \pi_{\text{city}}\bigl(\{\mathbf{s}_d(t)\}_{d=1}^{D},\; t\bigr)
\end{equation}

At each signalised intersection $l \in \mathcal{L}$, the cluster 
controller selects a signal action $\phi_l(t) \in \{\texttt{HOLD}, 
\texttt{SWITCH}\}$ at every simulation step $\Delta t$. 
The cluster controller $\pi_c$ combines the city-level directive 
$\mathbf{a}_d(t)$ with the local intersection state $\mathbf{s}_l(t)$ 
to produce a signal action:
\begin{equation}
    \phi_l(t) = \pi_c\bigl(\mathbf{s}_l(t),\; \mathbf{a}_d(t)\bigr), 
    \quad \forall\, l \in c
\end{equation}
where $\mathbf{s}_l(t)$ includes the current phase, time-in-phase 
$\tau_l(t)$, queue length, gap since last vehicle, and downstream 
occupancy at intersection $l$. The time spent in the current phase 
is subject to minimum and maximum green time constraints:
\begin{equation}
    g^{\min}_l \leq \tau_l(t) \leq g^{\max}_l
\end{equation}

\noindent where $g^{\min}_l$ and $g^{\max}_l$ are the minimum and maximum 
green time for intersection $l$.


\section{Hierarchical LLM Framework}
\label{sec:architecture}

\begin{figure*}[htb]
    \centering
    \includegraphics[width=\textwidth]{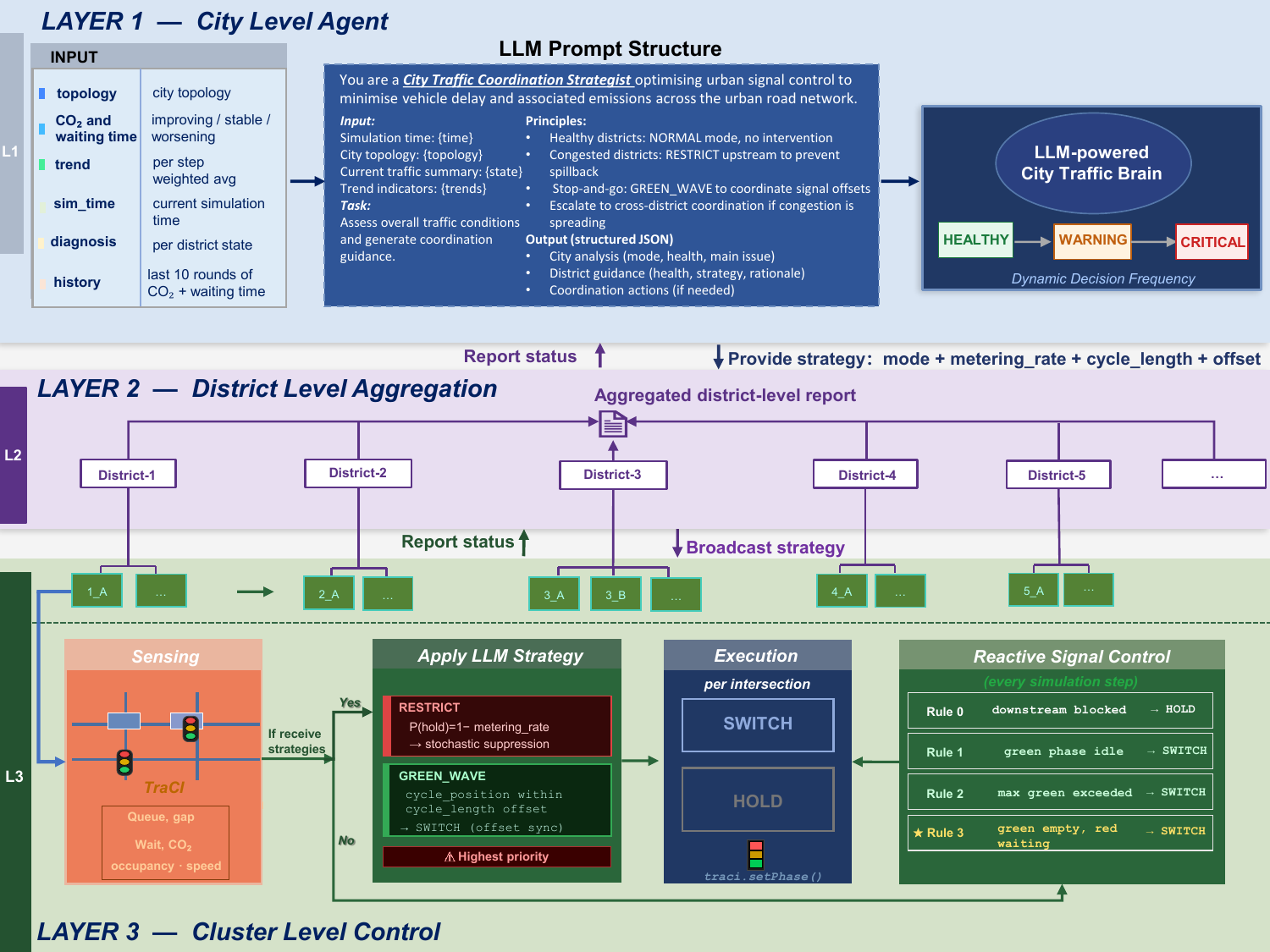}
    \vspace{-0.25in}
    \caption{Three-layer LLM-coordinated traffic signal control framework.}
    \vspace{-0.2in}
    \label{fig:framework}
\end{figure*}

\subsection{Framework Overview}

The proposed framework comprises three hierarchical layers shown in Fig.~\ref{fig:framework}: a city-level LLM agent issues strategic directives based on aggregated network conditions; a district-level layer distributes these to relevant clusters; and cluster-level controllers prioritize active directives, falling back to actuated logic otherwise. Information flows bidirectionally, closing the loop between global reasoning and local execution.

\subsection{City-Level Agent}

The city-level agent $\pi_{\text{city}}$ is responsible for 
monitoring city-wide traffic conditions and issuing coordination 
directives to all districts at dynamic decision interval.
At each invocation, the agent follows a three-step process. First, 
raw metrics reported from the cluster level (per-district 
CO\textsubscript{2} and weighted average waiting time) are processed 
internally to derive two additional inputs: a \textit{diagnosis} 
label per district (e.g., \texttt{IDLING\_CONGESTION}, 
\texttt{STOP\_AND\_GO}, \texttt{HEALTHY}) obtained by thresholding 
both metrics, and \textit{trend indicators} classifying each district 
as \texttt{improving}, \texttt{stable}, or \texttt{worsening} by 
comparing against the previous interval. Second, these inputs are 
assembled with the city topology and simulation time into a structured 
prompt, which is passed to the LLM via API. Third, the LLM returns 
a structured JSON response specifying, for each district, a 
coordination mode $m_d \in \{\texttt{NORMAL}, \texttt{RESTRICT}, 
\texttt{GREEN\_WAVE}\}$, a metering rate $\rho_d$, a target cycle 
length $T_d$, and a phase offset $\delta_d$, along with any 
cross-district coordination actions. 

\textit{Prompt Engineering:}
The city-level agent adopts a structured zero-shot prompt \cite{brown2020language} that contains no input-output demonstrations of the coordination task, as illustrated in Fig.~\ref{fig:framework}. The prompt consists of four sections: role and objective, current district state, trend indicators, and coordination principles. Role prompting frames the LLM as a 
\textit{City Traffic Coordination Strategist} with the explicit 
objective of minimising vehicle delay and associated emissions 
across the urban road network. The prompt enforces a strict JSON 
output schema with enumerated field values. The coordination principles instruct the LLM to assign one of three modes to each district via \texttt{district\_guidance}:

\begin{itemize}
    \item \texttt{NORMAL}: assigned to healthy districts with 
    $\rho_d = 1.0$, imposing no intervention. Cluster controllers 
    operate purely on local actuated logic.
    \item \texttt{RESTRICT}: assigned to a district exhibiting 
    congestion or high emission, applying intra-district metering 
    with $\rho_d \in (0,1)$ proportional to congestion severity.
    \item \texttt{GREEN\_WAVE}: issued when the agent identifies 
    potential for progressive vehicle flow along connected clusters 
    within a district, specifying a shared cycle length $T$ and 
    offsets $\{\delta_1, \delta_2, \ldots\}$ to allow platoons to 
    progress through consecutive intersections without stopping \cite{little1966synchronization}.
\end{itemize}

When congestion risk is detected as spreading across district 
boundaries, the LLM escalates to cross-district coordination by 
issuing additional actions via the \texttt{coordination} array:

\begin{itemize}
    \item \texttt{SPILLBACK\_CONTROL}: triggered when a downstream 
    district $\mathcal{D}_d$ is congested and worsening. The LLM 
    identifies the upstream district $\mathcal{D}_u$ feeding into 
    $\mathcal{D}_d$ and assigns $m_d = \texttt{RESTRICT}$ with 
    $\rho_d \in (0,1)$, reducing inflow to prevent spillback 
    propagation.
    \item \texttt{GREEN\_WAVE} (cross-district): issued when a 
    corridor spans multiple districts, specifying an ordered list 
    of cluster identifiers $\{c_1, c_2, \ldots\}$ with a shared 
    cycle length $T$ and offsets $\{\delta_1, \delta_2, \ldots\}$ 
    to coordinate platoon progression across district boundaries.
\end{itemize}


\subsection{District-Level Aggregation}
The district layer serves as an intermediary between global strategy and local execution. At each city-agent invocation, cluster-level metrics are aggregated into a district-level report that captures both typical load and tail congestion conditions, along with total CO$_2$ emissions. This aggregation is designed to ensure that localised hotspots within a district are not obscured by averaging across clusters. The resulting reports are passed to the city agent, and upon receiving coordination directives, the district layer broadcasts the corresponding parameters to all constituent clusters.

\subsection{Cluster-Level Control}

At cluster level, each controller $\pi_c$ manages a group of  intersections and 
executes one decision per simulation step $\Delta t$. The base logic is a 
reactive signal control policy, upon which city-level strategy is overlaid 
with higher priority.

\subsubsection{Reactive Signal Control}
At each step, $\pi_c$ evaluates four ordered rules per intersection:
\begin{itemize}
    \item Downstream blockage: Hold if downstream segment 
    occupancy exceeds a threshold.
    
    \item Gap detection: Switch if the measured headway on 
    the green approach exceeds
    \begin{equation}
        h_{\max} = \min\!\left(h^{\text{upper}},\,\max\!\left(h^{\text{lower}},\,
        \left(t_r + \frac{v}{2a}\right)\alpha\right)\right)
    \end{equation}
    where $h_{\max}$ is the headway threshold, $h^{\text{upper}}$ and 
    $h^{\text{lower}}$ are its upper and lower bounds, $t_r$ is driver 
    reaction time, $v$ is mean approach speed, $a$ is assumed 
    deceleration, and $\alpha$ is a calibration factor~\cite{Gazis1960}.
    
    \item Maximum green exceeded: Switch if the current phase 
    duration reaches $T^{\max}$ while queues remain on red-phase 
    approaches, preventing indefinite green extension under 
    sustained demand.
    
    \item Green-empty override: Switch if all lanes on the 
    current green phase are empty while one or more red-phase 
    approaches have waiting vehicles, avoiding unnecessary green 
    time on a cleared approach.
\end{itemize}

\subsubsection{Strategy Overlay}
City-level directives override the base policy before rule evaluation:

\begin{itemize}
    \item RESTRICT: Each switch decision is stochastically suppressed: with probability 
$(1-\rho_u)$ the phase is held regardless of gap or queue state, reducing cluster throughput 
by factor $\rho_u$.
    \item GREEN\_WAVE: When the cycle position falls within the switching window 
$[\delta_c - \epsilon,\,\delta_c + \epsilon]$ and the minimum green constraint is satisfied, 
a coordinated switch is triggered to align with the assigned platoon arrival pattern.
\end{itemize}

\section{Experimental Setup}
\label{sec:experiments}
Vehicle routes are derived from the SCATS traffic dataset \cite{ITSC}, data collected from 480 sensors in Dublin, the largest city in Ireland. Vehicle departure routes are sampled from this dataset to reflect realistic origin-destination patterns.

\subsection{Simulation Environment}

\begin{figure}[htbp]
    \centering
    \includegraphics[width=1.0\linewidth]{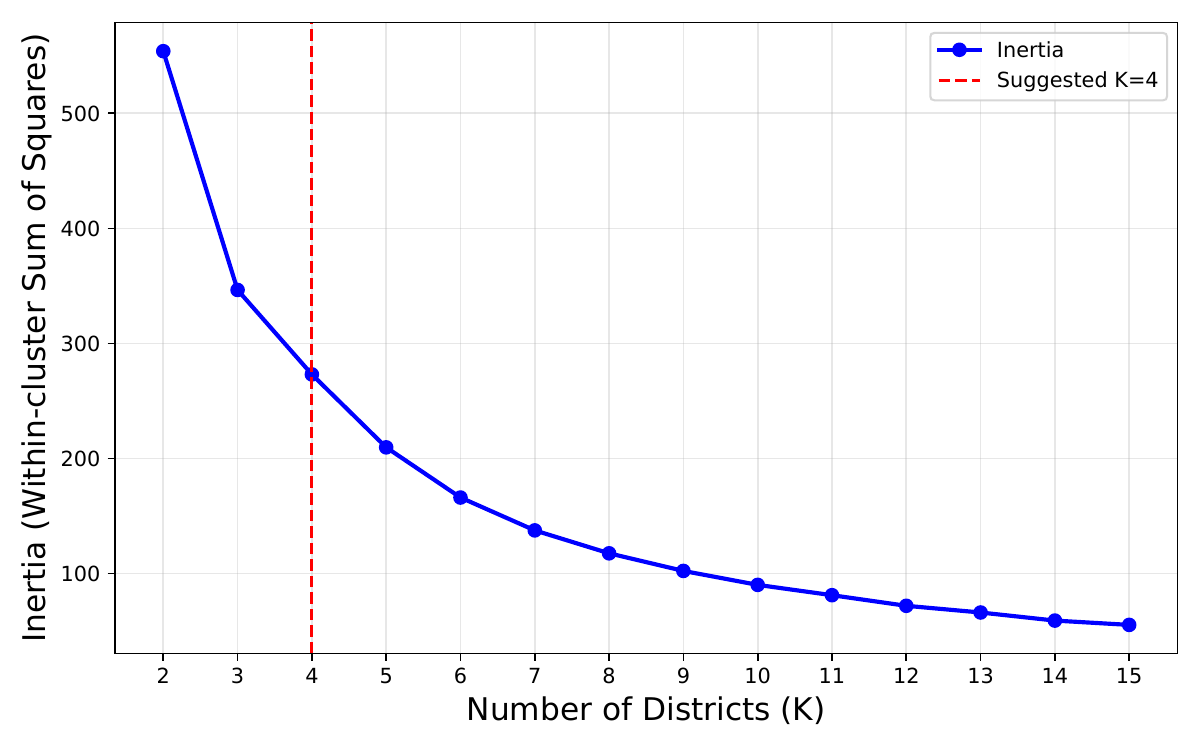}
    \vspace{-0.3in}
    \caption{Elbow Method on the Dublin traffic network.}
    \vspace{-0.1in}
    \label{fig:elbow}
\end{figure}

All experiments are conducted using SUMO on the DCC network, 
comprising 434 signalised intersections. To partition the network into manageable control regions, we apply K-Means clustering to the geographic coordinates of all signalised junctions. The optimal number of districts is determined via the Elbow Method \cite{thorndike1953belongs}: K-Means is run for $K = 2, \ldots, 15$,  the second-order difference method identifies $K=4$ as the geometric elbow (Fig.~\ref{fig:elbow}). However, since a 4-district partition places an excessive computational burden on each regional cluster controller given the density of the DCC network, we adopt $K=5$ as a practical compromise, yielding five districts and 176 signal clusters. The simulation step length is set to 0.5\,s. Two congestion levels are evaluated: a low-congestion scenario targeting approximately 1,200 vehicles in the network at steady state, and a high-congestion scenario with a total demand of 30,000 vehicles targeting 3,000 simultaneous vehicles. Each experiment is repeated with fixed random seeds for reproducibility.
\vspace{-0.15in}


\subsubsection{Vehicle Types} We model two vehicle classes using the HBEFA3 emission classification standard: passenger cars comprising 80\% of traffic, assigned the emission class \texttt{HBEFA3/PC\_G\_EU4}, and heavy-duty vehicles comprising the remaining 20\%, assigned \texttt{HBEFA3/HDV\_D\_EU4}. Both types follow the Intelligent Driver Model for car-following behavior. Traffic demand is composed of background vehicles, which maintain the steady-state congestion level in the simulation, and test vehicles, which are injected after network equilibrium is reached and serve as the evaluation subjects.

\subsubsection{Baselines}
We compare HiLLTS with three baselines:
\begin{itemize}
\item \textbf{Fixed-Time} control \cite{webster1958traffic} uses pre-timed signal plans that remain 
static throughout the simulation. The signal plans are automatically generated 
by SUMO based on the OpenStreetMap\footnote{https://www.openstreetmap.org/} road network topology.
\item \textbf{Actuated} control \cite{darroch1964queues} extends the green phase based on real-time gap detection, representing a strong and practically deployed baseline.
\item \textbf{Max Pressure} control~\cite{varaiya2013max} is a well-established adaptive strategy that maximises network throughput by balancing queue pressures across competing movements.
\end{itemize}

\subsubsection{LLM Configurations}
HiLLTS is evaluated across three model variants, namely Gemini-2.0-Flash, Gemini-3-Flash, and GPT-5 nano, to assess model performance under low and high congestion scenarios with respect to the choice of LLM. 


\subsubsection{LLM Parameters}
All LLMs are queried with the temperature set to 0 to ensure deterministic outputs. To balance computational cost and responsiveness, the city-level agent operates at a dynamic invocation interval adjusted based on the observed network state:
\[
\resizebox{\columnwidth}{!}{$
\Delta t_{\text{city}} = \begin{cases}
\Delta t_{\text{critical}} & \text{if } \bar{w} > \theta_{\text{critical}} \text{ or city health} = \texttt{CRITICAL}\\
\Delta t_{\text{urgent}}   & \text{if } \bar{w} > \theta_{\text{urgent}}   \text{ or city health} = \texttt{WARNING}\\
\Delta t_{\text{normal}}   & \text{otherwise}
\end{cases}
$}
\]
where $\bar{w}$ is the current city-wide average waiting time. In our 
implementation, $\theta_{\text{critical}} = 50$\,s, $\theta_{\text{urgent}} 
= 30$\,s, and $\Delta t_{\text{critical}} = 10$\,s, $\Delta t_{\text{urgent}} 
= 15$\,s, $\Delta t_{\text{normal}} = 60$\,s. These thresholds are inspired by the Highway Capacity Manual (HCM) \cite{hcm2016} LOS criteria, where LOS~D and LOS~E boundaries are defined at $35$\,s and $55$\,s; the conservative values of $30$\,s and $50$\,s are adopted to trigger responses before oversaturation onset \cite{roess2011traffic}. The $10$--$15$\,s update interval during congestion aligns with the phase-adjustment granularity of SCATS \cite{lowrie1982scats}, while the $60$\,s interval reflects the rolling optimization horizon of OPAC \cite{gartner1983opac}. This allows the LLM to respond 
rapidly during congested periods, while reducing API calls under 
stable conditions.

\begin{table*}[ht!]
\centering
\caption{Performance comparison. The best results are shown in bold. Underlined values indicate the second-best performance.}
\vspace{-0.1in}


\begin{tabular}{ c c c c c c c c}
\toprule
\multirow{2}{*}{\textbf{Scenario}} & 
\multirow{2}{*}{\textbf{Metric}} & 
\multicolumn{6}{c}{\textbf{Model}} \\ 
\cline{3-8}\\
& & Fixed-Time & Actuated & Max Pressure & \makecell{HiLLTS \\ (Gemini-2.0-Flash)} & 
\makecell{HiLLTS \\ (Gemini-3-Flash)} & 
\makecell{HiLLTS \\ (GPT-5 nano)} \\
\hline
\\
\multirow{3}{*}{Low congestion} 
& ACE {$\downarrow$}      & 1.00 & 0.91 & 0.89 & 0.87   & \textbf{0.82} & \underline{0.83} \\ 
& AWT{$\downarrow$}  & 148.04 & 106.49 & 88.76  & 71.72    & \textbf{56.16}  & \underline{67.93}  \\ 
& ATT & 460.02 & 411.00 & 394.84 & 372.63   & \textbf{357.44} & \underline{364.25} \\
\hline
 \\
\multirow{3}{*}{High congestion} 
& ACE {$\downarrow$}      & 2.89  & 2.45  & 3.15  & 2.42  & \textbf{2.24}  & \underline{2.32}  \\ 
& AWT {$\downarrow$}  & 1215.46 & 953.79  & 1363.90 & 923.40  & \textbf{813.46}  & \underline{884.50}  \\ 
& ATT {$\downarrow$}& 1674.62 & 1379.05 & 1832.41 & 1371.11 & \textbf{1258.34} & \underline{1320.96} \\
\bottomrule\\
\end{tabular}
\vspace{-0.1in}
\label{performance}
\end{table*}

\begin{figure*}[ht!]
    \centering
    \includegraphics[width=\textwidth]{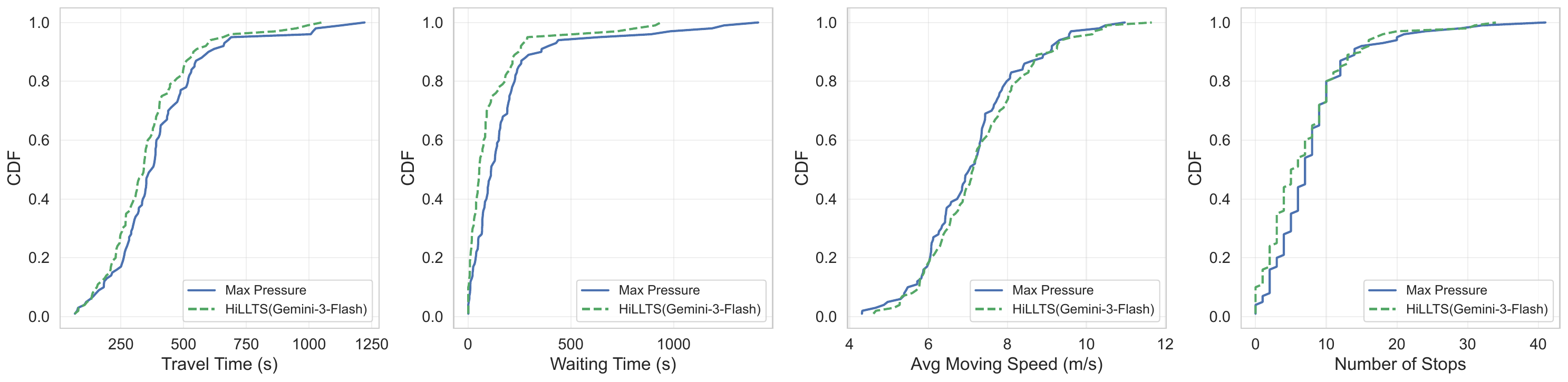}
    \vspace{-0.3in}
    \caption{CDF of per-vehicle travel metrics under low-congestion conditions. HiLLTS (Gemini-3-Flash) outperforms the Max Pressure baseline across all four metrics, with the most pronounced gains in waiting time and average moving speed.}
    \vspace{-0.1in}
    \label{fig:low}
\end{figure*}

\begin{figure*}[ht!]
    \centering
    \includegraphics[width=\textwidth]{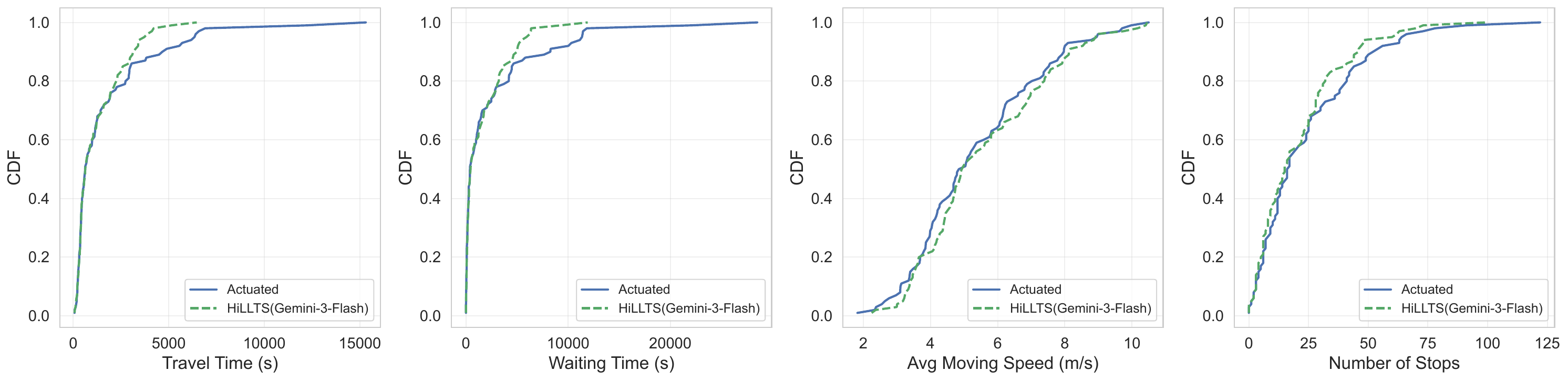}
    \vspace{-0.3in}
    \caption{CDF of per-vehicle travel metrics under high-congestion conditions. HiLLTS (Gemini-3-Flash) substantially reduces extreme travel and waiting times compared to the Actuated baseline.}
    \vspace{-0.3in}
    \label{fig:high}
\end{figure*}

\subsubsection{Evaluation Metrics}
All metrics are measured exclusively over the 100 test vehicles 
injected after the background traffic has reached a stable level, and background traffic serves solely to establish realistic congestion levels without confounding the evaluation.
\begin{itemize}
    \item \textbf{Average CO\textsubscript{2} Emissions (ACE):} Average cumulative CO\textsubscript{2} emissions (kg) produced by test vehicles over the entire simulation, computed using the HBEFA3 emission model embedded in SUMO. 
    \item \textbf{Average Waiting Time (AWT):} Average cumulative time (s) 
    spent stationary across all test vehicles throughout 
    their entire routes. This metric captures both signal-induced delay and 
    congestion-induced queuing, as a direct measure of overall 
    intersection-level control quality.
    \item \textbf{Average Travel Time (ATT):} Average end-to-end travel time(s) of all test vehicles from the origin to the destination, which includes both waiting time and any propagated delays due to congestion throughout the network. 
\end{itemize}

\section{Results}
\label{sec:results}
In this section, we present the experimental results under different congestion levels, along with an ablation study, to demonstrate the effectiveness of HiLLTS.

\subsection{Low Congestion Scenario}
In the low-congestion setting, all three variants of the HiLLTS model consistently outperform all 
baselines on all three metrics, as shown in Table~\ref{performance}. The Gemini-3-Flash variant yields the strongest overall performance, 
achieving the largest reductions across all baselines, most notably 18.00\% in ACE, 
62.07\% in AWT, and 22.30\% in ATT relative to Fixed-Time 
control, and 9.89\%, 47.26\%, and 13.03\% over Actuated control. The Gemini-2.0-Flash 
variant achieves reductions of 13.00\% in ACE, 51.56\% in AWT, 
and 19.00\% in ATT compared to Fixed-Time control. Against the Actuated 
baseline, improvements of 4.40\%, 32.65\%, and 9.34\% are observed for ACE, AWT, 
and ATT, respectively. Gains over Max Pressure further confirm the 
advantage of the proposed architecture over classical adaptive control. The GPT-5 nano variant 
also improves on all baselines, with reductions of 17.00\%, 54.12\%, and 20.82\% in ACE, 
AWT, and ATT versus Fixed-Time.

\subsubsection{Computational Overhead Under Low Congestion}

Under low congestion, the City Agent (powered by \textit{Gemini-3-Flash}) 
was invoked 73 times over a 1720.0\,s simulation, yielding a 
mean invocation interval of 23.6\,s. This interval falls 
between the 15\,s and 60\,s
thresholds, indicating moderate traffic fluctuations throughout the run. 
The mean API response time of approximately 3.5\,s per invocation is 
negligible relative to the city-level decision interval, 
and well within the phase durations of typical urban signal cycles, suggesting that LLM-guided strategic control may be feasible in simulation, though real-world deployment would require asynchronous execution and robust fallback control. The  API cost for the low-congestion scenario was \$0.62 USD (73 invocations, mean 2000 input and 2500 output tokens per call). 

\subsubsection{Per-Vehicle Variability Analysis}
To assess how HiLLTS affects individual vehicle experience under 
low-congestion conditions, we analyze the per-vehicle Cumulative 
Distribution Function (CDF) distributions of travel time, waiting 
time, average moving speed, and stop count in Fig.~\ref{fig:low}. 
Despite the relatively uncongested network, HiLLTS yields consistent 
improvements over the best baseline, Max Pressure, across all four 
metrics. The 95th-percentile waiting time decreases from 648\,s 
under Max Pressure to 299\,s under HiLLTS, a reduction of 53.86\%, 
and the 95th-percentile travel time decreases from 705\,s to 659\,s. 
Average moving speed is marginally higher and per-vehicle stop counts 
are reduced under HiLLTS. These results suggest that HiLLTS improves individual vehicle 
efficiency even under light network load.
\vspace{-0.05in}

\subsection{High Congestion Scenario}
In the high-congestion setting, all three HiLLTS variants consistently outperform all baselines 
again across all metrics (Table~\ref{performance}). Gemini-3-Flash achieves the largest 
improvements, reducing ACE by 22.49\%, AWT by 33.07\%, and ATT by 24.86\% compared 
to Fixed-Time control. It further outperforms Max Pressure by 28.89\%, 40.36\%, and 
31.33\% on ACE, AWT, and ATT respectively, and improves over Actuated control by 8.57\%, 
14.71\%, and 8.75\%. GPT-5 nano ranks second, with reductions of 19.72\%, 27.23\%, 
and 21.12\% in ACE, AWT, and ATT versus Fixed-Time, and 26.35\%, 35.15\%, and 
27.91\% over Max Pressure. Gemini-2.0-Flash demonstrates more modest gains under high 
congestion: 16.26\%, 24.03\%, and 18.12\% in ACE, AWT, and ATT versus Fixed-Time, 
and exhibits marginal improvements over Actuated 
baselines.

\subsubsection{Computational Overhead Under High Congestion}

Under high congestion, the City Agent was invoked 586 times over a 
6698.5\,s simulation, yielding a mean invocation interval of 11.4\,s, 
close to the minimum permitted interval of 10\,s, reflecting sustained 
near-critical traffic states throughout the run. Compared to the low 
congestion scenario, this represents a 2.1 times increase in 
invocation frequency, demonstrating that the dynamic frequency 
adaptation mechanism responds appropriately to deteriorating conditions. 
Each invocation produced on average 1.26 coordination actions, 
confirming that the architecture remains computationally feasible under 
demanding real-world conditions. The estimated API cost for the high-congestion scenario was approximately \$6.89 USD (586 invocations, $\sim$2,500 input / 3,500 output tokens per call).

\subsubsection{Per-Vehicle Variability Analysis}
To assess how HiLLTS affects the experience of individual vehicles in high congestion 
conditions, we analyze the CDF distributions per-vehicle of the four metrics in
Fig.~\ref{fig:high}. HiLLTS and the best baseline, Actuated, exhibit similar CDF 
curves in the lower percentiles, but diverge substantially in the upper tail. 
HiLLTS reduces the 95th-percentile travel time from approximately 15{,}000\,s 
to below 7{,}000\,s and compresses the worst-case waiting time by over 47\%, 
while also lowering the upper tail of per-vehicle stop counts. These results suggest that HiLLTS primarily suppresses severe congestion episodes rather than improving typical-case performance.

\subsection{Statistical Significance}

Within each simulation run, paired comparisons of 100 
test vehicles show consistent distributional differences between 
HiLLTS (Gemini-3-Flash) and all baselines (paired t-test, $p < 0.05$, n = 100 
vehicles). As all vehicles share the same network and control 
policy within a run, these results should be interpreted as 
indicative rather than inferential across independent trials. Effect sizes range from small to large (Cohen's $d = 0.208$--$1.463$), with the strongest effects observed under low congestion against Fixed-Time and Actuated baselines. Under high congestion, GPT-5 nano does not achieve statistical significance over Actuated ($p > 0.28$), highlighting the advantage of the stronger Gemini-3-Flash backbone in saturated network conditions. Table~\ref{tab:significance} summarises the $p$-values and effect sizes for the primary HiLLTS (Gemini-3-Flash) variant.

\vspace{-0.1in}
\begin{table}[h]
\centering
\caption{Statistical significance of HiLLTS (Gemini-3-Flash) improvements (paired t-test, $n=100$ vehicles).}
\vspace{-0.15in}
\label{tab:significance}
\renewcommand{\arraystretch}{1.1}
\begin{tabular}{llccc}
\toprule
\textbf{Scenario} & \textbf{Baseline} & \textbf{Metric} & \textbf{$p$-value} & \textbf{Cohen's $d$} \\
\midrule
\multirow{9}{*}{Low} 
  & \multirow{3}{*}{Fixed-Time}  & ATT & $<$0.001 & 0.876 \\
  &                               & AWT & $<$0.001 & 0.717 \\
  &                               & ACE & $<$0.001 & 0.479 \\
\cmidrule{2-5}
  & \multirow{3}{*}{Actuated}    & ATT & $<$0.001 & 1.463 \\
  &                               & AWT & $<$0.001 & 0.881 \\
  &                               & ACE & 0.001    & 0.332 \\
\cmidrule{2-5}
  & \multirow{3}{*}{Max Pressure} & ATT & $<$0.001 & 0.947 \\
  &                               & AWT & $<$0.001 & 0.472 \\
  &                               & ACE & 0.040    & 0.208 \\
\midrule
\multirow{9}{*}{High}
  & \multirow{3}{*}{Fixed-Time}  & ATT & $<$0.001 & 0.395 \\
  &                               & AWT & $<$0.001 & 0.384 \\
  &                               & ACE & $<$0.001 & 0.399 \\
\cmidrule{2-5}
  & \multirow{3}{*}{Actuated}    & ATT & 0.017    & 0.244 \\
  &                               & AWT & 0.013    & 0.254 \\
  &                               & ACE & 0.024    & 0.229 \\
\cmidrule{2-5}
  & \multirow{3}{*}{Max Pressure} & ATT & $<$0.001 & 0.614 \\
  &                               & AWT & $<$0.001 & 0.591 \\
  &                               & ACE & $<$0.001 & 0.582 \\
\bottomrule
\end{tabular}
\end{table}
\vspace{-0.2in}

\begin{table*}[ht!]
\centering
\caption{Ablation study on the district-level LLM agent in HiLLTS and rule-based control. Bold indicates best performance.}
\vspace{-0.1in}
\begin{tabular}{c c c c c}
\toprule
\multirow{2}{*}{\textbf{Scenario}} & 
\multirow{2}{*}{\textbf{Metric}} & 
\multicolumn{2}{c}{\textbf{HiLLTS (Gemini-3-Flash)}} & 
\multirow{2}{*}{\textbf{Rule-based}} \\ 
\cmidrule{3-4}
 & & City level with LLM only & City + District level with LLM & \\
\midrule
\multirow{3}{*}{Low congestion} 
 & ACE $\downarrow$      & \textbf{0.82} & 1.01 & 0.89 \\
 & AWT $\downarrow$ & \textbf{56.16}  & 81.94  & 79.57  \\
 & ATT $\downarrow$& \textbf{357.44} & 387.62 & 385.00 \\
\midrule
\multirow{3}{*}{High congestion} 
 & ACE $\downarrow$      & \textbf{2.24} & 3.13    & 2.46  \\
 & AWT $\downarrow$ & \textbf{813.46} & 1057.05   & 959.13  \\
 & ATT $\downarrow$& \textbf{1258.34}& 1510.74   & 1409.60 \\
\bottomrule
\end{tabular}
\vspace{-0.2in}
\label{tab:ablation}
\end{table*}

\subsection{Ablation Study}

To test whether adding an LLM agent at the district level provides 
additional benefit over the proposed architecture, and to quantify the overall contribution of LLM-guided coordination, we conduct an 
ablation study across three configurations, with results shown in 
Table~\ref{tab:ablation}: \textit{City LLM only} (proposed), 
\textit{City + District LLM} (with an additional district-level 
agent), and \textit{Rule-based} (no LLM baseline).
The \textit{City LLM only} configuration achieves the best performance 
across all metrics. Adding a district-level LLM not only fails to 
improve over the city-only architecture but also degrades performance below 
the rule-based baseline, increasing CO$_2$, waiting time, and trip 
duration under both scenarios. This validates the design 
choice of a single city-level LLM agent.


\section{Conclusion}
\label{sec:conclusion}

This work presents HiLLTS, a hierarchical LLM-guided traffic signal control framework in which a city-level LLM agent issues strategic 
coordination directives to constrained local cluster controllers, 
without requiring task-specific training or network-specific retraining. Evaluated on a SUMO network derived from DCC, the framework reduces average waiting time by 36.73\% and CO\textsubscript{2} 
emissions by 7.87\% under low congestion, and by 14.71\% and 8.57\% 
respectively under high congestion, compared with the strongest 
non-LLM baseline in each scenario. The results suggest that 
LLM-guided strategic coordination can improve delay and 
SUMO-estimated CO\textsubscript{2} emissions in the evaluated 
simulation scenarios. Future work should validate the framework 
across multiple calibrated networks, broader demand conditions, 
and stronger adaptive and learning-based baselines, and should 
address real-world signal constraints and pedestrian experience.

\bibliographystyle{IEEEtran}
\bibliography{IEEEexample}

@inproceedings{llmlight2025,
  author    = {Siqi Lai and Zhao Xu and Weijia Zhang and Hao Liu and Hui Xiong},
  title     = {{LLMLight}: Large Language Models as Traffic Signal Control Agents},
  booktitle = {Proceedings of the 31st ACM SIGKDD Conference on Knowledge Discovery
               and Data Mining (KDD)},
  year      = {2025},
  address   = {Toronto, Canada},
  publisher = {ACM},
  doi       = {10.5281/zenodo.14619359}
}

@article{llmrl2025,
  title={Real-Time Traffic Flow Optimization Using Large Language Models and Reinforcement Learning for Smart Urban Mobility},
  author={Singh, Arvind R and Ashraf, Muhammad Wasim Abbas and Rathore, Rajkumar Singh and Li, Bin and Sujatha, MS},
  journal={Applied Soft Computing},
  pages={113917},
  year={2025},
  publisher={Elsevier}
}

@inproceedings{yan2025llm,
  title={LLM-TrafficBrain: An Information-Centric Framework for Dynamic Signal Control with Large Language Models},
  author={Yan, Jiayu and Li, Donghe and Yang, Qingyu},
  booktitle={2025 IEEE 26th China Conference on System Simulation Technology and its Applications (CCSSTA)},
  pages={252--256},
  year={2025},
  organization={IEEE}
}

@article{vehicles2025,
  title={Large language models (llms) as traffic control systems at urban intersections: A new paradigm},
  author={Masri, Sari and Ashqar, Huthaifa I and Elhenawy, Mohammed},
  journal={Vehicles},
  volume={7},
  number={1},
  pages={11},
  year={2025},
  publisher={MDPI}
}

@article{tits2025crossroads,
  author    = {Mohammad Movahedi and Juyeong Choi},
  title     = {The Crossroads of {LLM} and Traffic Control: A Study on Large
               Language Models in Adaptive Traffic Signal Control},
  journal   = {IEEE Transactions on Intelligent Transportation Systems},
  volume    = {26},
  pages     = {1701--1716},
  year      = {2025},
  publisher = {IEEE},
  doi       = {10.1109/TITS.2024.3498735}
}

@inproceedings{icca2025,
  title={LLM-Enhanced MARL for Smarter Traffic Control},
  author={Chen, Xingmei and Meng, Wei},
  booktitle={2025 IEEE 19th International Conference on Control \& Automation (ICCA)},
  pages={535--540},
  year={2025},
  organization={IEEE}
}

@article{hybridtrafficai2025,
  author    = {Hazrat Bilal and Abbas Rehman and Muhammad Shamrooz Aslam and
               Inam Ullah and Wen-Jer Chang and Neeraj Kumar},
  title     = {Hybrid {TrafficAI}: A Generative {AI} Framework for Real-Time
               Traffic Simulation and Adaptive Behavior Modeling},
  journal   = {IEEE Transactions on Intelligent Transportation Systems},
  year      = {2025},
  publisher = {IEEE},
  doi       = {10.1109/TITS.2025.11018863}
}

@inproceedings{matsim2025,
  title={Bridging ai and traffic simulation: A robust framework for llm-based ai replanning agents in matsim},
  author={Patwary, Ashraf Uz Zaman and Ciari, Francesco and Angioloni, Luca and Naseri, Hamed and Brusci, Lorenzo and Iannelli, Giulio and Bakhtiari, Arsham},
  year={2024},
  publisher={Bureau de Montreal, Universit{\'e} de Montreal}
}

@misc{survey2025,
  author        = {Yimo Yan and Yejia Liao and Guanhao Xu and Ruili Yao and
                   Huiying Fan and Jingran Sun and Xia Wang and Jonathan Sprinkle and
                   Ziyan An and Meiyi Ma and Xi Cheng and Tong Liu and Zemian Ke and
                   Bo Zou and Matthew Barth and Yong-Hong Kuo},
  title         = {Large Language Models for Traffic and Transportation Research:
                   Methodologies, State of the Art, and Future Opportunities},
  year          = {2025},
  eprint        = {2503.21330},
  archivePrefix = {arXiv},
  primaryClass  = {cs.CE}
}

@INPROCEEDINGS{ITSC,
  author={Guériau, Maxime and Dusparic, Ivana},
  booktitle={2020 IEEE 23rd International Conference on Intelligent Transportation Systems (ITSC)}, 
  title={Quantifying the impact of connected and autonomous vehicles on traffic efficiency and safety in mixed traffic}, 
  year={2020},
  volume={},
  number={},
  pages={1-8},
  doi={10.1109/ITSC45102.2020.9294174}}

@article{thorndike1953belongs,
  title={Who belongs in the family?},
  author={Thorndike, Robert L},
  journal={Psychometrika},
  volume={18},
  number={4},
  pages={267--276},
  year={1953},
  publisher={Springer-Verlag}
}

@article{varaiya2013max,
  title={Max pressure control of a network of signalized intersections},
  author={Varaiya, Pravin},
  journal={Transportation Research Part C: Emerging Technologies},
  volume={36},
  pages={177--195},
  year={2013},
  publisher={Elsevier}
}

@article{blondel2008fast,
  title={Fast unfolding of communities in large networks},
  author={Blondel, Vincent D and Guillaume, Jean-Loup and Lambiotte, Renaud and Lefebvre, Etienne},
  journal={Journal of statistical mechanics: theory and experiment},
  volume={2008},
  number={10},
  pages={P10008},
  year={2008}
}

@article{Gazis1960,
  author  = {Gazis, Denos C. and Herman, Robert and Maradudin, Alexei},
  title   = {The problem of the amber signal light in traffic flow},
  journal = {Operations Research},
  year    = {1960},
  volume  = {8},
  number  = {1},
  pages   = {112--132},
  doi     = {10.1287/opre.8.1.112}
}

@inproceedings{mcqueen1967some,
  title={Some methods of classification and analysis of multivariate observations},
  author={McQueen, James B},
  booktitle={Proc. of 5th Berkeley Symposium on Math. Stat. and Prob.},
  pages={281--297},
  year={1967}
}

@article{khreis2017health,
  title={Health impacts of urban transport policy measures: A guidance note for practice},
  author={Khreis, Haneen and May, Anthony D and Nieuwenhuijsen, Mark J},
  journal={Journal of Transport \& Health},
  volume={6},
  pages={209--227},
  year={2017},
  publisher={Elsevier}
}

@article{eom2020traffic,
  title={The traffic signal control problem for intersections: a review},
  author={Eom, Myungeun and Kim, Byung-In},
  journal={European transport research review},
  volume={12},
  number={1},
  pages={50},
  year={2020},
  publisher={Springer}
}

@article{chu2019multi,
  title={Multi-agent deep reinforcement learning for large-scale traffic signal control},
  author={Chu, Tianshu and Wang, Jie and Codec{\`a}, Lara and Li, Zhaojian},
  journal={IEEE transactions on intelligent transportation systems},
  volume={21},
  number={3},
  pages={1086--1095},
  year={2019},
  publisher={IEEE}
}

@article{choi2025optimizing,
  title={Optimizing Traffic Signal Control Using LLM-Driven Reward Weight Adjustment in Reinforcement Learning},
  author={Choi, Sujeong and Lim, Yujin},
  journal={Journal of Information Processing Systems},
  volume={21},
  number={1},
  pages={43--51},
  year={2025},
  publisher={Korea Information Processing Society}
}

@article{wang2023carbon,
  title={Carbon dioxide emission reduction-oriented optimal control of traffic signals in mixed traffic flow based on deep reinforcement learning},
  author={Wang, Zhaowei and Xu, Le and Ma, Jianxiao},
  journal={Sustainability},
  volume={15},
  number={24},
  pages={16564},
  year={2023},
  publisher={MDPI}
}

@article{anthi2025role,
  title={The role of artificial intelligence in shaping intelligent motorways: opportunities, challenges, and real-world implementations},
  author={Anthi, Eirini and Williams, Lowri and Afzal, Hamza Ahmad and Brar, Bilal Ahmad and Bhowmick, Joydip and Gujral, Kabir and Thomas, Emyr},
  journal={IEEE Transactions on Intelligent Transportation Systems},
  year={2025},
  publisher={IEEE}
}

@book{hcm2016,
  author    = {{Transportation Research Board}},
  title     = {Highway Capacity Manual 6th Edition},
  year      = {2016},
  publisher = {TRB}
}

@book{roess2011traffic,
  author    = {Roess, R. P. and Prassas, E. S. and McShane, W. R.},
  title     = {Traffic Engineering},
  year      = {2011},
  publisher = {Pearson}
}

@article{lowrie1982scats,
  author    = {Lowrie, P. R.},
  title     = {The {Sydney} Co-ordinated Adaptive Traffic System: Principles, Methodology, Algorithms},
  journal   = {IEE Conf. Road Traffic Signalling},
  year      = {1982}
}

@article{gartner1983opac,
  author    = {Gartner, N. H.},
  title     = {{OPAC}: A Demand-Responsive Strategy for Traffic Signal Control},
  journal   = {Transportation Research Record},
  volume    = {906},
  year      = {1983}
}

@techreport{webster1958traffic,
  title={Traffic signal settings},
  author={Webster, Fo Vo},
  year={1958}
}

@article{brown2020language,
  title={Language models are few-shot learners},
  author={Brown, Tom and Mann, Benjamin and Ryder, Nick and Subbiah, Melanie and Kaplan, Jared D and Dhariwal, Prafulla and Neelakantan, Arvind and Shyam, Pranav and Sastry, Girish and Askell, Amanda and others},
  journal={Advances in neural information processing systems},
  volume={33},
  pages={1877--1901},
  year={2020}
}

@article{little1966synchronization,
  title={The synchronization of traffic signals by mixed-integer linear programming},
  author={Little, John DC},
  journal={Operations Research},
  volume={14},
  number={4},
  pages={568--594},
  year={1966},
  publisher={Informs}
}

@article{darroch1964queues,
  title={Queues for a vehicle-actuated traffic light},
    author = {Darroch,  J. N. and Newell,  G. F. and Morris,  R. W. J.},
  journal={Operations Research},
  volume={12},
  number={6},
  pages={882--895},
  year={1964},
  publisher={INFORMS}
}

\end{document}